\documentclass[sn-nature,iicol,Numbered]{sn-jnl}

\usepackage{graphicx}%
\usepackage{multirow}%
\usepackage{amsmath,amssymb,amsfonts}%
\usepackage{amsthm}%
\usepackage{mathrsfs}%
\usepackage[title]{appendix}%
\usepackage{xcolor}%
\usepackage{textcomp}%
\usepackage{manyfoot}%
\usepackage{booktabs}%
\usepackage{algorithm}%
\usepackage{algorithmicx}%
\usepackage{algpseudocode}%
\usepackage{listings}%

\theoremstyle{thmstyleone}%
\theoremstyle{thmstyletwo}%

\theoremstyle{thmstylethree}%

\begin{document}

\title[Article Title]{Can CNNs Accurately Classify Human Emotions? A Deep-Learning Facial Expression Recognition Study}


\author*[1]{\fnm{Ashley} \sur{Jisue Hong}}\email{ahpuppy63@gmail.com}

\author[2]{\fnm{David} \sur{DiStefano}}\email{david.distefano@tufts.edu}

\author[3]{\fnm{Sejal} \sur{Dua}}\email{sejaldua@gatech.edu}

\affil*[1]{ \orgname{Punahou School}, \orgaddress{\street{1601 Punahou St.}, \city{Honolulu}, \postcode{96813}, \state{HI}, \country{USA}}}

\affil[2]{\orgdiv{Psychology}, \orgname{Tufts University}, \orgaddress{\street{490 Boston Ave.}, \city{Medford}, \postcode{02155}, \state{MA}, \country{USA}}}

\affil[3]{\orgdiv{Data Science}, \orgname{Georgia Institute of Technology}, \orgaddress{\street{225 North Ave.}, \city{Atlanta}, \postcode{30332}, \state{GA}, \country{USA}}}


\abstract{Emotional Artificial Intelligences are currently one of the most anticipated developments of AI. If successful, these AIs will be classified as one of the most complex, intelligent nonhuman entities as they will possess sentience, the primary factor that distinguishes living humans and mechanical machines. For AIs to be classified as “emotional,” they should be able to empathize with others and classify their emotions because without such abilities they cannot normally interact with humans. This study investigates the CNN model’s ability to recognize and classify human facial expressions (positive, neutral, negative). The CNN model made for this study is programmed in Python and trained with preprocessed data from the Chicago Face Database. The model is intentionally designed with less complexity to further investigate its ability. We hypothesized that the model will perform better than chance (33.3\%) in classifying each emotion class of input data. The model accuracy was tested with novel images. Accuracy was summarized in a percentage report, comparative plot, and confusion matrix. Results of this study supported the hypothesis as the model had 75\% accuracy over 10,000 images (data), highlighting the possibility of AIs that accurately analyze human emotions and the prospect of viable Emotional AIs.
}

\keywords{Robotics and Intelligent Machines, Machine Learning, Convolutional Neural Network (CNN), Human Emotion}



\maketitle

\section{Introduction}\label{sec1}

Artificial Intelligence systems, most commonly referred to as AI systems, are computer-based systems or machines that perform tasks that have traditionally been thought to require human intelligence or higher order thinking. AI systems are most commonly used to help solve human problems and better procedures \cite{Tai2020-oh}. Because AI systems are developed to imitate human cognition, scientists constantly compare the limits and abilities of an AI system to those of a human. Currently, studies still find that humans are better than AI systems in solving broader problems under various circumstances, especially in social interactions \cite{Korteling2021-fk}. However, unlike simple machines that can only perform one task without further development, AI systems can constantly evolve and learn by themselves to expand their limits by developing their own decision-making process based on the given information. The subset of processes by which an AI develops its own way of thinking is called machine learning. According to the National Library of Medicine, machine learning or deep learning refers to the various algorithms that generate educated predictions based on a large dataset \cite{Nichols2019-vq}. 

Artificial neural networks are a popular type of machine learning algorithm. According to the National Library of Medicine, artificial neural networks are developed based on the biological concept of human neurons \cite{Han2018-wt}. Within each artificial neural network are multiple nodes, which each act like a human neuron. Each node is connected to corresponding layers that help pass on and assess the initial message. A typical artificial neural network contains an input layer, various nodes, a couple of hidden layers, and an output layer. When an input is given, each node connected to the input layer uses its individually assigned weight and threshold to assess the input’s average numeric values, which represents how close the input is to the stored data for classification. When the output (input’s average numeric value) of any singular node is above the threshold value, it activates the node to pass on its output to the next layer, and the process continues until it reaches the final output node and layer. If the input of a node is below the threshold, it will not continue any process and halt. 

Among the various kinds of neural networks are convolutional neural networks, also known as CNNs. Convolutional neural networks are the most popular neural network for computer vision and image analysis \cite{Yamashita2018-tu}. While they can be used in other applications like natural language processing as well, CNNs are mostly used for classification of images. Although looking at an object and being able to classify it is a basic human trait in vision, computers cannot differentiate simple objects before they go through complex deep learning like CNNs. With multiple hidden (convolutional) layers, pooling layers, and fully connected layers, CNNs allow AI to assess each image data with more accuracy and store more data. The reason why CNNs are most often used for image analysis is because traditional neural networks, also called multilayer perceptrons (MLP), are quite inefficient in their way of processing images. When MLP tries to process large images, they run out of storage space and lose crucial information about the input images. While MLPs struggle just to store an image in its network to process, CNNs quickly find patterns in an image and assess the classification based on the features it detects by itself through their convolutional layers that each assess a different feature of the image and altogether compile the closeness of an image to an existing class in the neural network’s realm of information. While the input data for MLPs are vectors, CNNs accept matrices as input and compile them into tensors, containers of numeric data, expanding the number of hidden layers and increasing the range of patterns it can detect. By having multiple layers, the CNN will have a larger compile of hidden layers (larger tensor), which means that it is “deeper” in terms of tensor dimensions. Because CNNs have a more complex structure with more advanced mathematical components, they are able to process images with higher accuracy.

Emotional AI systems are machines that can recognize and classify human emotions. While current AI systems have already exhibited the potential to surpass the limits of human intelligence, they are still not the equivalent to humans because AIs lack the ability to feel and understand human emotions.  If successful, emotional AIs will be regarded the most similar to human minds and be considered the most intelligent form of AI. As AIs become more and more sophisticated and intelligent compared to real human beings, a key question that determines the relationship between human beings and AI systems emerges: can AIs ever be considered sentient? In 2022, an engineer working for Google, Blake Lemoine, was placed on paid administrative leave after arguing that Google’s AI chatbot LaMDA is sentient, showing signs of being “alive.” One of the key examples Lemoine used to support his argument was how LaMDA said it feels its own emotions, such as “pleasure, joy, love, sadness, depression, contentment, anger, and many others.” From this it can be inferred that LaMDA could be considered sentient if it is validated as a functioning emotional AI. 

In order for Lemoine’s theory to be accepted as true, LaMDA not only has to show the ability to generate its own emotions but also prove that it has emotional intelligence, commonly described as empathy or the ability to understand others’ emotions. According to a journal-published study by neuroscientist Aron K. Barbey et. al., in 2012, scientists proved that “emotional intelligence and cognitive intelligence share many neural systems for integrating cognitive, social, and affective processes,” so “there are interdependencies between emotional intelligence and general intelligence” \cite{Barbey2014-si}. An important factor of emotional intelligence is empathy, a feeling that unites every well human being. According to the European Journal of Personality, psychopaths are people who have shallow emotions and lack empathy. In relation to these facts, according to the European Journal of Personality, psychopaths scored significantly lower on intelligence tests when tested against normal humans without any mental illnesses \cite{Sanchez_de_Ribera2019-bb}. Oftentimes, psychopaths are distinguished from others because their lack of empathy and understanding in human emotions cause them to only think for themselves, preventing them from normally interacting with others. It is therefore crucial for an emotional AI to have emotional intelligence since they will be classified more similar to socially isolated psychopaths than normal humans without it.  

One possible way for an AI to gain emotional intelligence and show empathy is to be able to assess and classify others’ emotions. To assess the possibility of emotional AIs and learn how well AIs can recognize emotions, this study investigates the accuracy of a CNN model in classifying human facial expressions. This study investigates the classification of human emotions through facial expression exhibited within the Chicago Face Database where pictures of individuals may display a positive, neutral, or negative expression \cite{Ma2015-no}. The model CNN is programmed with Python and PyTorch. Based on other similar research, we hypothesize that this model will perform better than chance (33.3\%) in classifying each class of given emotions based on the input data. This study’s results will not only be helpful in determining the possibility of developing an emotional AI but will also be useful in developing technology that simply requires AIs to read facial expressions. Example uses of this study’s results are managing workplace and employee satisfaction, student stress monitoring, remote therapy and guidance, etc \cite{Mantello2023}.

Over the last few years, studies on AIs and machine learning models that recognize facial expressions have increased. Though there are some previous studies on the same subject, this study differs with them for various reasons. Unlike how this study aims to investigate the possibility of an Emotional AI and the accessibility of CNN-based Facial Expression Recognition Systems (FER), the study published in 2017 by V. Tümen aimed to diversify computer vision applications through building a CNN based FER \cite{Tumen2017-ao}. Tümen’s model was not trained with CFD but was trained with the popular FER2013 database, and had a 57.1\% success rate \cite{Lasri2019-ls}. The study published in 2019 by I. Larsi took this subject at another angle and elaborated how CNNs can be used to analyze students’ emotions, but like Tümen’s study they too used the FER2013 dataset \cite{Lasri2019-ls}. Some other differences between this study and Larsi’s was that they kept the RGB color factor of the pictures for their model to process, had a smaller pixel size for the input compared to this study, had more emotion classes than the three this study has, and thus had a different model architecture \cite{Lasri2019-ls}. One major difference between the studies of Tümen and Larsi was their success rate. While they used the same dataset, their results varied by a lot in percentage, with Tümen’s model producing a 57.1\% accuracy while Larsi’s producing a 70.1\% accuracy \cite{Tumen2017-ao}\cite{Lasri2019-ls}. To clarify and investigate if CNN-based FERs are actually viable, this study uses CFD’s Expression Morph dataset besides the popular FER2013 dataset commonly used for CNN-based FERs. To mitigate racial bias during training, this study uses grayscale to normalize all images (data), and uses a simpler model architecture to match the smaller size of the dataset.  

\section{Methods}

\subsection{Dataset}

The dataset used in this study was provided by the Chicago Face Database developed by Debbie S. Ma, Joshua Correll, and Bernd Wittenbrink of the University of Chicago \cite{Ma2015-no}. The specific dataset used in study is the “Expression Morphs” set based on the original CFD dataset developed by Vikhanova, Mareschal, and Tibber Queen Mary University of London \cite{Vikhanova2022-yy}. The data contained in the datasets were 320 headshots of multiple human individuals with varying facial expressions per individual. The individuals in the data varied in sex (male and female), race (Black and White), and age (unspecified), but all headshots were taken with the same white background and all individuals wore the same gray shirt.

\subsection{Data Processing}

\begin{figure}[hbt!]%
\centering
\includegraphics[width=0.9\linewidth]{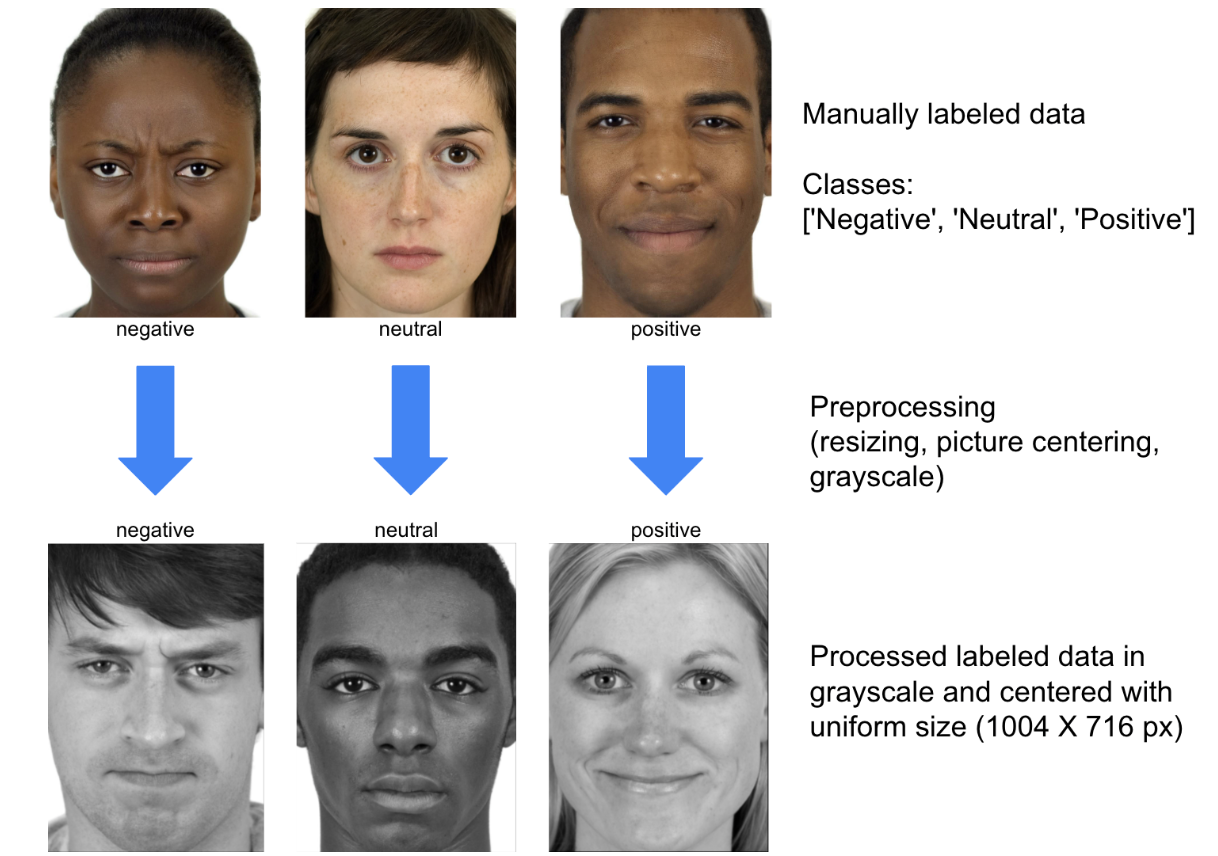}
\caption{Image processing procedure including resizing and scaling.}
\label{fig1}
\end{figure}

Data was preprocessed before it was used for training the CNN model. Figure \ref{fig1} shows the procedure of data processing with examples of the data. Firstly, all headshots were manually labeled as displaying “positive,” “negative,” or “neutral” expressions before they were used to train the model. All data manually sorted from the CFD datasets with labeled emotions were uploaded to the Google Drive folder connected to the model and each emotion label was set as classes. Then the data was processed to all have the same pixel size because the layers of the model require a fixed input size to produce comparable outputs. During resizing the headshots were also centered in order to focus on the facial features that display human expression. Finally, the data was all processed in grayscale to mitigate possible racial bias and reduce another factor for the model to practice so that the model runs with better efficiency.

\subsection{Experimental Design}

\begin{figure}[hbt!]%
\centering
\includegraphics[width=0.9\linewidth]{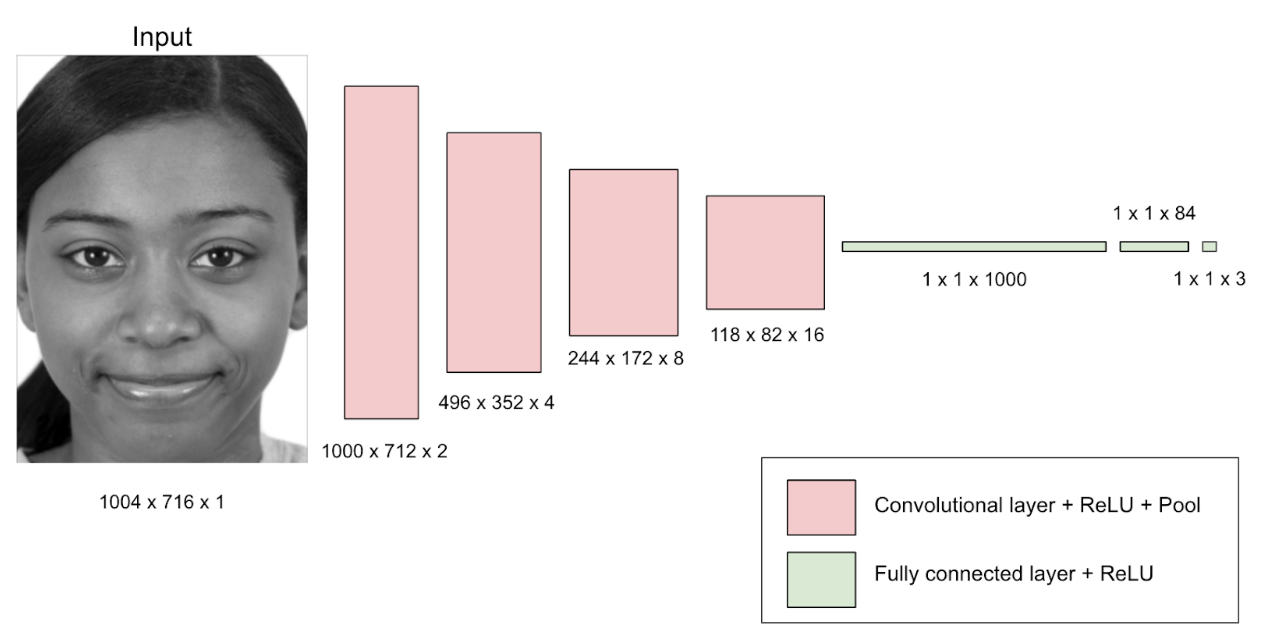}
\caption{Model architecture diagram.}
\label{fig2}
\end{figure}

The model CNN is designed with various architectures for the best performance. Figure \ref{fig2} shows the model architecture and each layer’s dimensions. The model is built with four convolutional layers and three fully connected layers to amplify its accuracy. The images are separated into two data groups, train data and test data. The dataset went through full randomization when it was split into train data and test data. Because it was fully randomized, the train and test data have a chance of having data (headshot) of the same individual. However, since the two groups of data were completely randomized the expression that the individual displays in train and test data is different and there is no headshot that exists in both the train data group and test data group. The train data is 75\% (240 pictures), and the test data is 25\% (80 pictures) of the total data.

After the model was trained with the train data over 100 epochs, the model went through two tests to assess its accuracy. First it iterated over the entire train data again. Second, the model was tested with novel images (a.k.a. the test data) to investigate if it is intelligent enough to assess new data. The model’s performance was evaluated using three methods: classification accuracy (\%), confusion matrix, and graphing test accuracy over 100 epochs.

\section{Results}

\begin{figure}[hbt!]%
\centering
\includegraphics[width=0.9\linewidth]{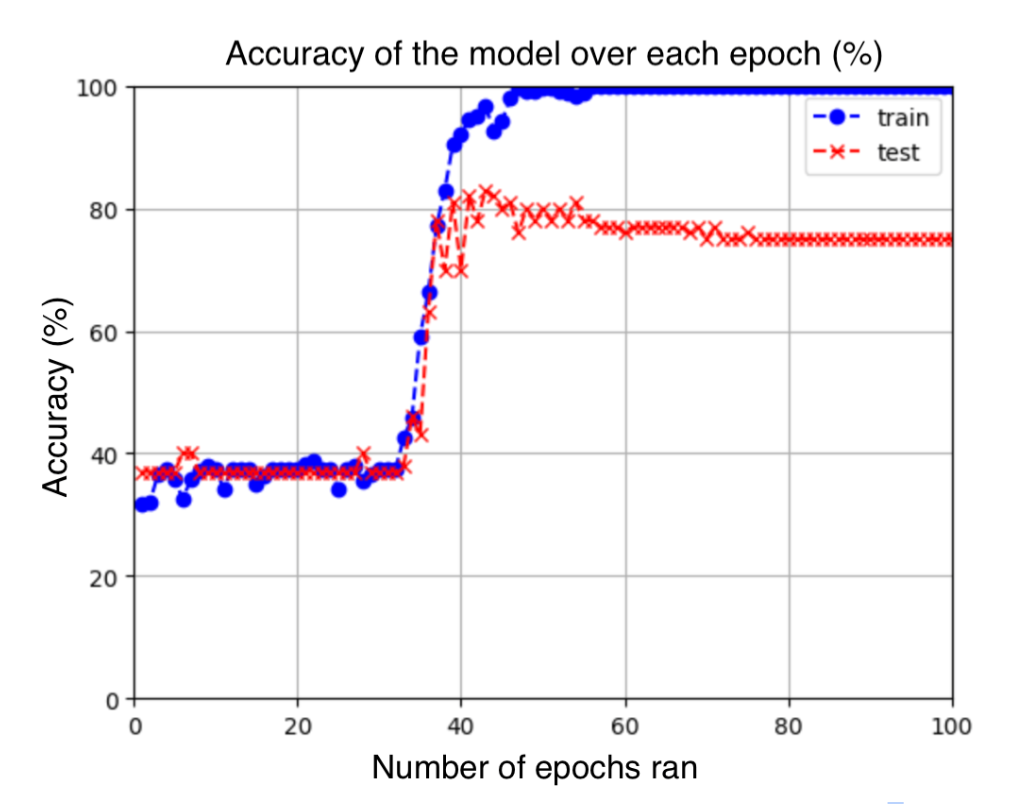}
\caption{Graph of the model’s accuracy (\%) history over each epoch run.}
\label{fig3}
\end{figure}

When commanded to calculate the model’s accuracy, the code printed that the accuracy of the network on the 10,000 test images is 75\%. Figure \ref{fig3} displays how the model’s accuracy reaches for both the train and test data were similar to chance (33.3\%) for the first 35 epochs, but after the 35th epoch the training accuracy reaches 100\% and the validation (testing) accuracy reaches 75\%. 

\begin{figure}[hbt!]%
\centering
\includegraphics[width=0.9\linewidth]{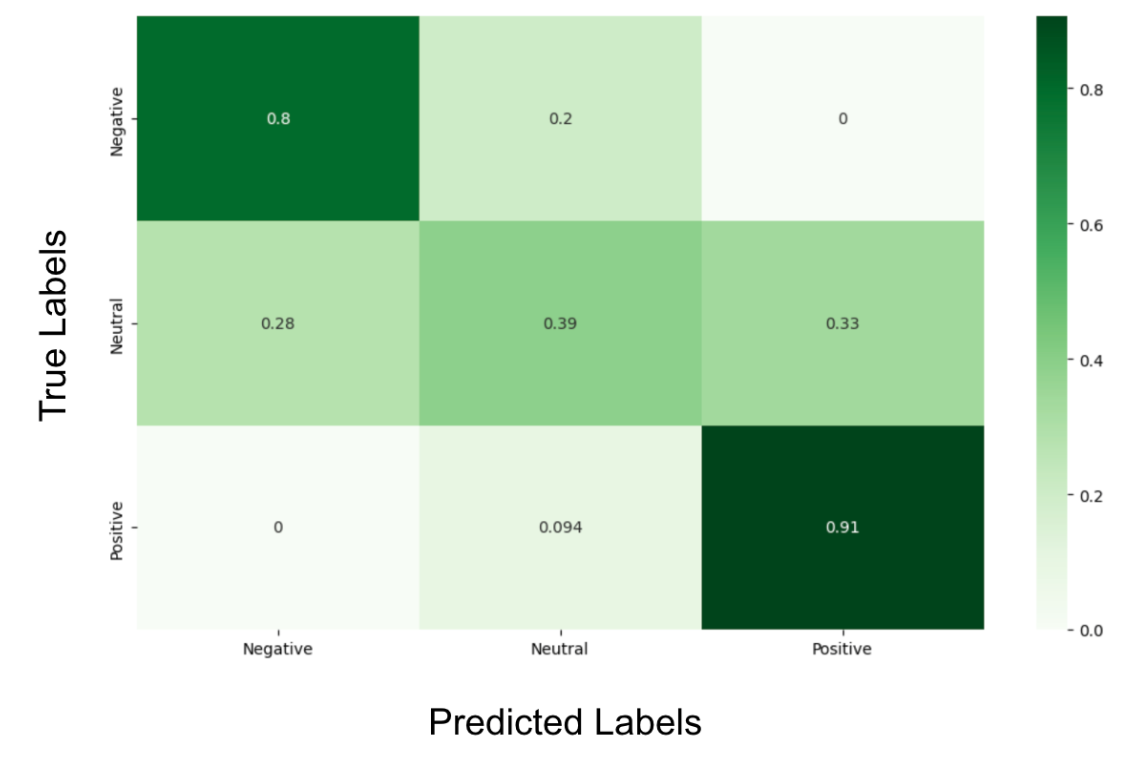}
\caption{Confusion matrix illustrating types of misclassified observations.}
\label{fig4}
\end{figure}

Figure \ref{fig4} shows the model’s testing accuracy on predicting each emotion class through a confusion matrix. On Figure 4, the vertical column represents true labels and the horizontal row represents predicted labels. Each centered number on the boxes of the matrix represents the percentage of data in each true emotion class to be classified as a certain predicted emotion class by the model. As shown in the rightmost color legend, the darker the color of the box the higher the probability is for that box. For example, the upper leftmost box with the true label “Negative” and predicted label “Negative” shows that about 80\% of data with the true class “Negative” is classified “Negative” by the model. Figure \ref{fig4} shows that the model had about 80\% accuracy in classifying “Negative” data, 39\% accuracy in classifying “Neutral” data, and 91\% accuracy in classifying “Positive” data. In conclusion, the results supported the hypothesis. 

\section{Discussion}

The results supported the hypothesis, with the model producing an overall accuracy rate of 75\% even with a limited dataset and less layers. There are many factors which lead up to this achievement of the model. Starting with the positive factors, the model’s performance met the expectations, and was higher than Tümen’s study with the conventional FER2013 dataset.10 As shown in Figure \ref{fig3}, the model had varying accuracy between the three emotion classes. The model showed significantly high accuracy in classifying data with classes “Negative” and “Positive,” each with about 80\% accuracy and 91\% accuracy. In contrast, the model did not show high accuracy in classifying “Neutral” data, with about 39\% accuracy that is slightly higher than chance (33.3\%). A possible explanation of this result could be that while facial features displaying negative and positive expressions are more distinguishable and similar, neutral expressions of individuals can easily vary depending on their original facial features such as existing wrinkles. 

One factor that might have allowed the model to perform better is the data normalization of colors. According to a study in 2012 by Klare et al., FERs often have a negative racial bias towards certain races, especially towards people of color \cite{klare}. Klare’s study highlights that their models showed the most bias towards Black females between the ages of 18 and 30 \cite{klare}. Though it may be important for the model to differentiate race for identification AIs, this study focused on assessing the model’s ability to recognize human expressions and emotions, so it was decided that the RGB factor would be removed from all data through normalization before training the model, thus having all data in grayscale. Besides eliminating possible racial bias, removing colors from the pictures has also allowed the model to run faster since there was one less component for it to process. One concern regarding racial bias is that CFD’s Expression Morph dataset did not have any data from races besides Black and White \cite{Vikhanova2022-yy}. This may cause the model to show racial bias against races other than Black and White if it encounters other novel data because even though all training data is normalized with grayscale, people from other races have physically different facial features (i.e. eye shape, mouth shape, etc.) that the model was not trained to recognize. In order to avoid such possible racial bias, more data of other races should be added to the training and testing data for future models. 

Another factor that could have influenced the model’s performance is the small dataset and minimal layers in its architecture. The model’s architecture was purposefully designed minimally with only four convolutional layers and three fully connected layers to investigate if such CNN FER models can be simple enough to implement accessibly for various purposes. The architecture seemed to work well with the inputted dataset, and it ran through well over the 100 epochs. This showed that the minimalistic model architecture can be applicable for classification of other datasets in use cases with acceptable speed and accuracy. The size of the dataset had both pros and cons. While the small dataset with fewer classes allowed the model to run and train faster, it might have also caused the model to overfit over the training data. Evidence of this can be seen in Figure \ref{fig2}, where the epoch accuracy for the train data remains near 100\% starting from about 45 epochs run while the epoch accuracy for the test data stabilizes around 75\% starting from about 70 epochs run. Adding more data and classes for the model or reducing the number of epochs in training could possibly enhance the model’s accuracy by avoiding overfitting. 

The last factor to consider for future references is the amount of information the model has to process when it trains and tests through the inputted data. It is a possible concern that CFD and its Expression Morph dataset might have too many unnecessary components for the model to process, including hair, clothes, ears, and other components of the picture that do not reflect a person’s emotion through facial expression. It is also possible that those components irrelevant to the person’s facial expressions not only slow down the model but also create biases that tie those components with emotion labels. An example could be that a person’s ear shape may be considered by the model in classifying her emotions. To make the model more efficient and accurate, it may be beneficial to process the data even more before training so that the model only focuses on facial components (i.e. eyes, wrinkles, eyebrows, mouth, etc.) that actually display the emotions. 

\section{Conclusion}

In conclusion, the CNN model meets the expectations that provide the possibility of emotional AIs and how well they can recognize emotions. There are definitely some areas that can be better supported with more data within this study, but the model has shown potential in accuracy and efficiency with minimal architecture and an unconventional dataset, adding support to the conclusion that CNN FERs can be pretty accurate. If more improvements are implemented, it is possible that the model can show greater accuracy that introduces more feasibility in creating an Emotional AI in the future.

\section*{Acknowledgments}

I thank David DiStefano for his project mentoring, Sejal Dua for her showcasing mentoring, and the CFD Database for providing the data for this project. 

\bibliographystyle{chem-acs}
\bibliography{manuscript}

\end{document}